\documentclass[10pt, a4paper]{article}

\usepackage[final]{lrec2026} % this is the new style
\usepackage{hyperref}
\usepackage{float}
\usepackage{verbatim} %comments
\usepackage{apalike}
\restylefloat{figure}
\floatstyle{plaintop} %table caption at top
\restylefloat{table}
\usepackage{amssymb}
\usepackage{amsmath}

\usepackage{times}
\usepackage{latexsym}
\usepackage[utf8]{inputenc}
\usepackage{microtype}
\usepackage{inconsolata}
\usepackage{graphicx}

\usepackage[utf8]{inputenc}
\usepackage[T1]{fontenc}

\usepackage[table,svgnames]{xcolor}
\usepackage{soul}
\usepackage{color}
\definecolor{lightgreen}{RGB}{194, 245, 186}
\definecolor{darkgreen}{RGB}{80, 182, 39}
\definecolor{lightblue}{RGB}{163, 223, 255}

\usepackage{multirow}
\usepackage{makecell}
\usepackage{arydshln}
\usepackage{adjustbox}

\usepackage{booktabs}

\usepackage{xcolor}
\usepackage{soul}

\usepackage{comment}

\usepackage{xspace}

\usepackage{cleveref}
\usepackage{natbib}

\usepackage{kotex}

\usepackage{listings}

\crefformat{section}{\S#2#1#3}
\crefformat{subsection}{\S#2#1#3}
\crefformat{subsubsection}{\S#2#1#3}

\usepackage{url}
\usepackage{subcaption}

\newcommand{\minisection}[1]{%
\vspace{0.04in}
    \noindent \textbf{#1}.\xspace%
}

\newcommand{\ours}{$\textsc{PivotE}$\xspace}

\newcommand{\nllb}{$\text{NLLB}$\xspace}

\newcommand{\blender}{$\text{LLM-Blender}$\xspace}

\newcommand{\fid}{$\text{FiD}$\xspace}

\newcommand{\trice}{$\text{TRICE}$\xspace}

\newcommand{\hlc}[2][yellow]{{%
    \colorlet{foo}{#1}%
    \sethlcolor{foo}\hl{#2}}%
}

\title{A Single Model Ensemble Framework for Neural Machine Translation using Pivot Translation}

\name{Seokjin Oh$^{1\dagger}$, Keonwoong Noh$^2$, Woohwan Jung$^{2*}$
\thanks{
        $^\dagger$ Work done while at Hanyang University.
    }
\thanks{
        $^*$ Corresponding author.
    }
}

\address{$^1$SK Siltron, $^2$Korea University \\
seokjin.oh@sk.com, \{nohkw011, woohwan\}@korea.ac.kr\\}

\abstract{
Despite the recent remarkable advances in neural machine translation, translation quality for low-resource language pairs remains subpar.
Ensembling multiple systems is a widely adopted technique to enhance performance, often accomplished by combining probability distributions.
However, previous approaches face the challenge of high computational costs for training multiple models.
Furthermore, for black-box models, averaging token-level probabilities at each decoding step is not feasible.
To address the problems of multi-model ensemble methods, we present a pivot-based single model ensemble.
The proposed strategy consists of two steps: pivot-based candidate generation and post-hoc aggregation.
In the first step, we generate candidates through pivot translation.
This can be achieved with only a single model and facilitates knowledge transfer from high-resource pivot languages, resulting in candidates that are not only diverse but also more accurate.
Next, in the aggregation step, we select $\textit{k}$ high-quality candidates from the generated candidates and merge them to generate a final translation that outperforms the existing candidates.
Our experimental results show that our method produces translations of superior quality by leveraging candidates from pivot translation to capture the subtle nuances of the source sentence.
\\ \newline  }

\begin{document}

\maketitleabstract

\section{Introduction}

Neural machine translation (NMT) models exhibit outstanding capabilities when a large volume of the parallel corpus is available (e.g., translating from and to English).
However, their performance still falls short in cases involving low-resource languages (e.g., Basque) and translating between non-English languages from different language families (e.g., German-Russian)~\cite{artetxe2018unsupervised}.
Top-performing large language models (LLMs), such as GPT models~\cite{gpt3.5}, also demonstrate suboptimal translation performance in low-resource language pairs~\cite{robinson-etal-2023-chatgpt, moslem-etal-2023-adaptive}. % zhu2023multilingual
The scarcity of parallel data, primarily due to limited cultural interaction, makes the low-resource translation task more challenging.

In many generation tasks, ensembling multiple systems has proven to be a successful strategy for performance enhancement.
In NMT, traditional ensemble methods average probability distributions over output tokens from multiple models during decoding.
However, the high cost of training multiple models is the primary shortcoming of ensemble decoding.
Additionally, computing token-level probabilities at each decoding step is not feasible with recent black-box models such as GPT-4o and Gemini~\cite{gpt4o, geminiteam2023gemini}.

Ensemble methods that can be utilized even when token-level probabilities cannot be computed have also been proposed.
A selection-based ensemble method involves generating candidates from multiple models and then selecting the best candidate among them~\cite{wang2022rationaleaugmented, howgood}.
However, in this ensemble fashion, the final output space is limited to the existing candidate pool.
In contrast, the generation-based ensemble, such as \blender~\cite{llm-blender}, creates improved outputs using candidates obtained from multiple models.
This approach aims to generate a final output superior to the existing candidates.
Nonetheless, the main drawback of the notably high cost of generating candidates through multiple models remains, inducing computational overhead.
As the size of the models used in the ensemble increases, the cost proportionally escalates, becoming more burdensome.
In addition, due to the varying performance of MT systems, the quality of some candidates can be significantly lower than that of others, leading to a degradation in the overall performance.

To alleviate the problems above, we propose \textbf{Pivot}-based single model \textbf{E}nsemble (\ours), a novel generation-based approach.
Our intuition of a single model ensemble primarily stems from pivot translation, which can produce diverse and more accurate translations.
Pivot translation~\cite{wu-wang-2007-pivot, utiyama-isahara-2007-comparison} is a method that splits the end task into two sequential steps: source$\rightarrow$pivot and pivot$\rightarrow$target.
Pivoting has been employed to enhance low-resource translation by transferring knowledge from high-resource pairs.
In many cases, English, being a resource-rich language, serves as the intermediate language.
However, we employ not only English but various pivot languages for candidate generation, thereby producing diverse hypotheses using a single model.

In the next aggregation step, we select the top candidates for the ensemble and merge them to generate the final output.
Since the quality of candidates directly impacts the results of the ensemble, it is important to select high-quality candidates.
Given that the best pivot language for translation varies with each source sentence, we select the top-$\textit{k}$ candidates for each source sentence via quality estimation (QE).
By leveraging diverse candidates from pivot translation and knowledge of the merging module, \ours generates final translations that accurately convey the meaning and subtle nuances of the source sentence, superior to selecting from pre-existing candidates.
Our contributions can be summarized as follows:

\begin{itemize}
    \item We propose a simple but effective pivot-based single model ensemble method, \ours, to improve low-resource MT.

    \item We show that a single model can effectively generate diverse and accurate hypotheses and that leveraging these candidates in an ensemble process can enhance translation quality while reducing computational overhead.

    \item The empirical results on various language pairs demonstrate that we consistently outperform state-of-the-art methods, validating the effectiveness of the pivot-based ensemble.
\end{itemize}
\section{Related Work}

\minisection{Pivot-based approaches}
Pivot translation is an approach that decomposes the translation task into two sequential steps~\cite{wu-wang-2007-pivot, utiyama-isahara-2007-comparison}.
By transferring knowledge from high-resource pivot languages, pivoting is especially effective in translation between low-resource languages \cite{zoph-etal-2016-transfer, aji-etal-2020-neural, he-etal-2022-tencent}.
In this study, pivot translation enables us to obtain high-quality candidates for the ensemble.
\citet{kim-etal-2019-pivot} discusses a pivot-based transfer learning technique where source$\rightarrow$pivot and pivot$\rightarrow$target models are first trained separately, then use pre-trained models to initialize the source$\rightarrow$target model, allowing effective training of a single, direct NMT model.
\citet{zhang-etal-2022-triangular} further investigates the transfer learning approach by utilizing auxiliary monolingual data.

Pivot translation typically employs English as the bridge language.
Nonetheless, previous studies have explored the use of diverse pivot languages, taking into account factors such as data size and the relationships between languages~\cite{paul2009importance, dabre-etal-2015-leveraging}.
By leveraging the ability of pivot translation to produce diverse outputs, several studies have focused on generating paraphrases~\cite{mallinson-etal-2017-paraphrasing, guo2019zeroshot}.
More recently, \citet{mohammadshahi-etal-2024-investigating} uses pivot translation for an ensemble, but it requires computing token-level probabilities and fails to improve translation.
Our work shares motivation with these studies, generating translations depending on the pivot path to obtain a variety of candidates.

\minisection{Ensemble in NLG tasks}
Ensemble learning is a widely adopted strategy to obtain more accurate predictions by employing multiple systems~\cite{sagi2018ensemble}.
In NMT, the traditional approach involves averaging the probability distributions of the next target token, which is predicted at each decoding step by multiple models~\cite{bojar-etal-2014-findings} or by different snapshots~\cite{huang2017snapshot}.
When multiple sources are available, an ensemble can be conducted with predictions obtained from different sources~\cite{firat-etal-2016-zero}.
A token-level ensemble through vocabulary alignment across LLMs has also been proposed~\cite{eva}.
However, these methods are not applicable to recent black-box models as they cannot compute token-level probabilities at decoding time.

A selection-based ensemble has also been explored, which chooses the final output among the existing candidates.
This can be achieved through majority voting by selecting the most frequent one~\cite{wang2022rationaleaugmented} or selecting the best candidate with QE~\cite{fernandes-etal-2022-quality, howgood}.
Recently, MBR decoding~\cite{GOEL2000115, mbr}, which aims to find the hypothesis with the highest expected utility, has gained attention.
However, this approach limits the final output space to the existing candidate pool.

\begin{figure*}[t]
  \centering
  \includegraphics[width=\textwidth]{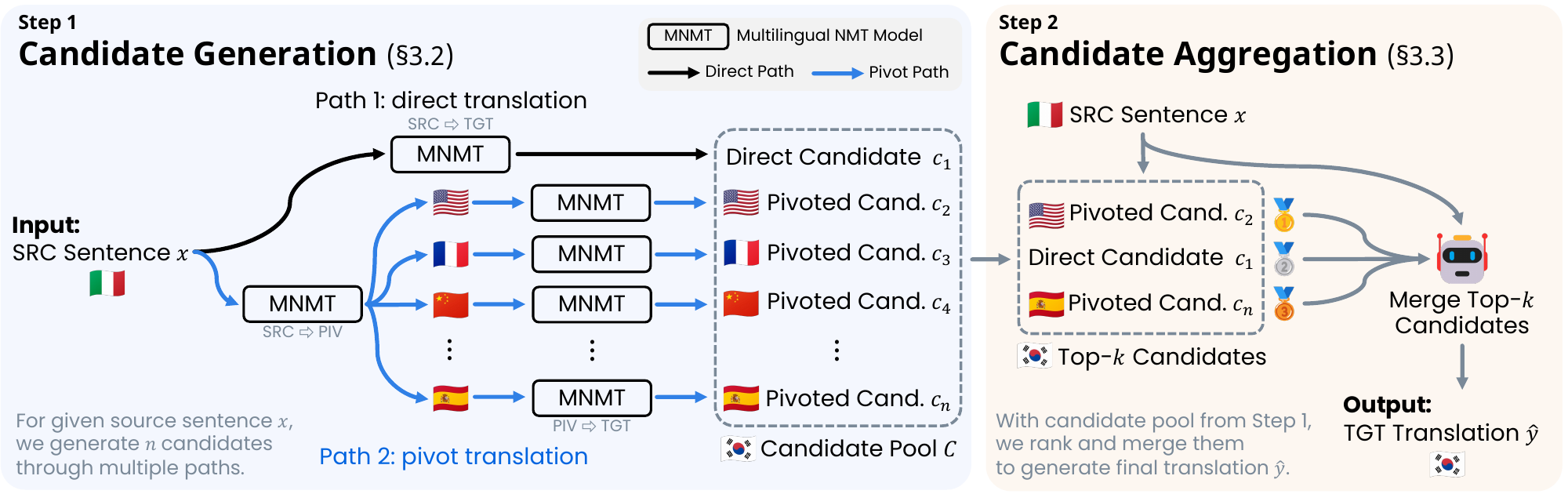} 
  \caption{Overview of \ours framework.}
  \label{fig:overall}
\end{figure*}

On the other hand, the generation-based ensemble method involves generating a new final prediction.
Fusion-in-Decoder~\cite{fid} proposes an architecture that aggregates additional information with a given input.
More recently, within the context of LLMs, \citet{llm-blender} and \citet{exchangeofthought} investigate a method of using LLMs to generate multiple outputs and aggregate them.
Generating new outputs through LLMs offers the benefit of explicitly harnessing their pre-trained knowledge within the ensemble process.
\section{Pivot-based Single Model Ensemble}
\label{sec:Pivot-based Single Model Ensemble}

In this section, we first introduce a overview of \ours framework (\S\ref{sec:overview}).
Then, we describe the candidate generation process through pivot translation (\S\ref{sec:pivot-based candidate generation}) and the aggregation process (\S\ref{sec:candidate aggregation}).

\subsection{Overview}
\label{sec:overview}

Our objective is the same as that of conventional translation tasks: converting the given source language sentence $x$ into a target language sentence $\hat{y}$.
\ours consists of two steps: candidate generation and candidate aggregation.
Figure~\ref{fig:overall} illustrates an overview of the proposed ensemble framework.

As the first step, we input $x$ to generate candidates through a single multilingual NMT model.
One translation path could be directly translating from the source to the target through the source$\rightarrow$target path.
Alternatively, pivot translations can be achieved by employing high-resource pivot languages, enabling translation paths of source$\rightarrow$pivot and pivot$\rightarrow$target.
During the pivot process, leveraging abundant parallel data enables knowledge transfer from high-resource pivot languages, thereby facilitating the generation of diverse and more accurate translations.
Through these $n$ paths, we can obtain a candidate pool $C = \{c_1, ..., c_n\}$ composed of $n$ candidates in the target language, employing only a single model.

As the second step, a ranking process is first conducted within the candidate pool $C$ since not all candidates contribute to the ensemble.
Using the estimated quality of each candidate, we select the top-$\textit{k}$ candidates.
We then generate the final output $\hat{y}$ using the selected high-quality candidates.
This generation-based approach facilitates the production of outputs superior to the existing candidates.

\subsection{Pivot-based Candidate Generation}
\label{sec:pivot-based candidate generation}

In the first step, \ours takes a source sentence $x$ as input and generates $n$ candidates.
Direct translation yields only one candidate, whereas pivot translation enables the generation of multiple candidates from a single source sentence using a single model.
Generating candidates through pivot translation has two major advantages: diversity and quality.

First, we can obtain diverse candidates that can act complementarily.
One of the key principles of the ensemble is that the participants must be sufficiently diverse to provide various inductive biases.
In \ours, each source sentence is translated diversely by passing through multiple translation paths.
Diverse translation paths enhance the likelihood of providing expressions that convey the accurate meaning of the source sentence.
Pivot-based candidate generation shares a similar goal with a previous study that generates paraphrases through round-trip translation, aiming to generate diverse translations~\cite{thompson-post-2020-paraphrase}.

Second, by utilizing a parallel corpus of high-resource pivot languages, pivoting enables more accurate translations.
For low-resource language pairs, more appropriate translations can be achieved through two-step decoding via a pivot language~\cite{he-etal-2022-tencent}.
Moreover, leveraging pivot languages with abundant parallel data, not limited to English, allows us to obtain better translations~\cite{paul2009importance, dabre-etal-2015-leveraging}.

In addition, pivot translation with a single model offers practical benefits over employing multiple models. 
Firstly, it can reduce the costs of operating multiple models, including LLMs. 
Secondly, the substantial performance disparities among models mean that using the top-performing single model for candidate generation often leads to higher-quality outcomes. 
Lastly, it reduces inference latency by using a single model for batched inferences, while multi-model ensembles require up to 11 models, causing significant overhead and limiting real-time response capability.
Given that pivot translation with a single model allows for the creation of diverse and more accurate translations, we utilize an MNMT model to generate the candidates.

\minisection{Selecting pivot languages}
For each language pair, we carefully select pivot languages based on the assumption that pivot languages with abundant mutual knowledge would allow us to obtain higher-quality candidates.
We select $n$ top-performing paths for our study based on BLEU scores on the FLORES-200 benchmark~\citeplanguageresource{nllb}.
We evaluate the outputs for each path, including direct translation and through various pivot translations.
\nllb~\cite{nllb} is used to generate candidates, and results on the FLORES-200 for selecting translation paths are in Appendix~\ref{sec:apdx_top4 pivot langauges}.
If pivot languages are selected based on BLEU scores, high-resource languages are predominantly chosen, rather than low-resource ones.
The experiments detailed in Appendix~\ref{apdx:resource level of pivot languages} demonstrate that overly prioritizing diversity by employing low-resource pivot languages, at the expense of candidate quality, does not result in improvements in the final translation.
The experiments comparing metrics for selecting translation paths are in Appendix~\ref{apdx: Metric for Selecting Translation Paths}.
As a result, we compose the candidate pool using the 4 paths.

\subsection{Candidate Aggregation}
\label{sec:candidate aggregation}

In the aggregation step, we take the candidate pool $C$ as input and output the merged final translation $\hat{y}$.
The post-hoc aggregation process encompasses two stages: selecting and merging.
In the first stage, we select candidates by a ranking method.
There are two approaches to selecting candidates.
One approach evaluates each translation path and selects the best paths for all source sentences.
The other approach involves selecting the best top-$\textit{k}$ candidates for each source sentence.
After selecting $\textit{k}$ candidates, we generate the final translation $\hat{y}$ using the merging module.
This process enables the creation of better outputs beyond the quality of existing candidates.

\minisection{Selecting the top-$\textit{k}$ candidates}
The pivot language that generates the highest-quality candidate varies for each source sentence.
The best output is not guaranteed from one translation path alone, as it can vary depending on factors such as the size of the parallel corpus and the relationship between languages.
First, \ours uses QE to rank all $n$ candidates from candidate pool $C = \{c_1, ..., c_n\}$.
Afterward, we select top-$\textit{k}$ candidates from $n$ candidate pool.
Selecting the top-$\textit{k}$ candidates ensures the quality of the output by filtering out low-quality candidates while also efficiently reducing the cost during the merging process.
We use the reference-free COMETkiwi (\textit{wmt22-COMETkiwi-da}) \cite{rei2022cometkiwi} for ranking candidates.

\minisection{Generating the final translation}
We generate the final translation $\hat{y}$ by merging the top-$\textit{k}$ candidates using two methods: LLM-based approaches and encoder-decoder ensemble architectures.

LLM-based approaches offer the advantage of implicitly leveraging their translation capabilities during the ensemble, as the source sentence is also provided.
We conduct experiments with \textsc{GenFuser}~\cite{llm-blender}, Llama-3~\cite{llama3modelcard}, and GPT models~\cite{gpt3.5, gpt4, gpt4o}. 
When employing \textsc{GenFuser}, we construct the input by concatenating the top-\textit{$k$} candidates to the prompt, as presented in \citet{llm-blender}.
For merging with Llama-3 and GPT, we use the ensemble prompt template in Appendix~\ref{sec:apdx_prompt_templates}.

On the other hand, encoder-decoder architectures employ smaller models dedicated to the ensemble, effectively reducing both training and inference costs.
We conduct experiments using Fusion-in-Decoder (FiD)~\cite{fid} and TRICE~\cite{trice}.
For FiD, the instruction and the source sentence are concatenated with each candidate and independently encoded.
The decoder then takes the concatenated representations and generates the final translation.
For TRICE, the model is trained with a two-stage fine-tuning method.
In the first stage, the model is trained on two different inputs and a single target: Source$\rightarrow$Target and Candidate$\rightarrow$Target.
In the second stage, the source and the candidate are concatenated and provided as a single input.
The overall input formats are:
\texttt{Translate source into <target language>}
\texttt{referring <target language> candidate.}
\texttt{source: <$x$>}
\texttt{candidate: <$c_{k}$>}
for FiD, and
\texttt{<$x$></$s$><$l_{s}$>;<$c_{1}$></$s$><$l_{t}$>;...;<$c_{k}$></$s$><$l_{t}$>}
with the language token \texttt{<$l_{lang}$>} for TRICE.
We provide architectural illustrations of these encoder-decoder ensemble architectures in Appendix~\ref{apdx:Fid/TRICE illustration}.

By leveraging various candidates, each with different strengths, the aggregation process can effectively mitigate errors in a complementary manner.
\section{Experiments}

We use NVIDIA RTX 3090 or 4090 GPUs for experiments.

\subsection{Datasets}
\label{sec:datasets}

\begin{table}[h!]
\centering
\begin{adjustbox}{width=\columnwidth, center}
\renewcommand{\arraystretch}{1.1}
\begin{tabular}{ccccc}
\Xhline{3\arrayrulewidth}
\multirow{2}{*}{\textbf{Lang-pair}} & \multirow{2}{*}{\textbf{Dataset}} & \multicolumn{3}{c}{\textbf{\# Sentences}} \\ \cline{3-5} 
 & & \textbf{Train} & \textbf{Dev} & \textbf{Test} \\ \hline\hline 
\multirow{2}{*}{KO $\leftrightarrow$ IT} & TED 2020 v1& \multirow{2}{*}{357,733} & \multirow{2}{*}{2,000} & \multirow{2}{*}{2,000} \\
 & \citeplanguageresource{ted2020} & & & \\ \hline  
\multirow{2}{*}{AR $\leftrightarrow$ PT} & WikiMatrix v1& \multirow{2}{*}{153,441} & \multirow{2}{*}{2,000} & \multirow{2}{*}{2,000} \\
 & \citeplanguageresource{schwenk2019wikimatrix}& & & \\
\Xhline{3\arrayrulewidth}
\end{tabular}
\end{adjustbox}
\caption{Dataset statistics.}
\label{tab:dataset statistics}
\end{table}

We conduct experiments on the linguistically distant languages within pairs: not in the same language family and using different scripts.
We select 2 language pairs, resulting in 4 translation directions in total, Korean (\textit{Koreanic})$\leftrightarrow$Italian (\textit{Romance}) and Arabic (\textit{Arabic})$\leftrightarrow$Portuguese (\textit{Romance}).
The language family grouping is defined by the criteria presented in \citet{m2m100}.

We validate our approach across various domains.
For Korean$\leftrightarrow$Italian pair, we run experiments on TED2020~\cite{ted2020}.
For Arabic$\leftrightarrow$Portuguese pair, we use WikiMatrix~\cite{schwenk2019wikimatrix}.
All the datasets are obtained from the OPUS\footnote{\url{https://opus.nlpl.eu}}~\citeplanguageresource{opus} project.
The statistics for the datasets are listed in Table~\ref{tab:dataset statistics}.

\begin{table*}[t]
\centering
\renewcommand{\arraystretch}{1.1}
\begin{adjustbox}{width=0.95\textwidth, center}
\begin{tabular}{lcccccccccccc}
\Xhline{3\arrayrulewidth}

\multirow{2}{*}{\textbf{Model}}  & \multicolumn{3}{c}{\textbf{Korean$\rightarrow$Italian}} & \multicolumn{3}{c}{\textbf{Italian$\rightarrow$Korean}} & \multicolumn{3}{c}{\textbf{Arabic$\rightarrow$Portuguese}} & \multicolumn{3}{c}{\textbf{Portuguese$\rightarrow$Arabic}}\\ \cline{2-13}
 & BLEU & chrF++ & COMET & BLEU & chrF++ & COMET  & BLEU & chrF++ & COMET & BLEU & chrF++ & COMET \\ \hline\hline 

\textit{\textbf{Standalone NMT System}}\\ \hdashline[3pt/3pt]
NLLB~\cite{nllb} & 16.27 & 41.14 & 84.60 &            17.40 &           23.39 &           87.33 &           27.25 &           50.35 &           84.21 &            13.50 &            40.90 &           84.24 \\
Vicuna~\cite{Vicuna} &10.11  & 31.15 & 70.29 & 10.60 & 16.51 & 72.29 & 17.64 & 38.44 & 76.01 & 8.40 &  27.38& 79.18 \\
Baize~\cite{baize} & 10.62 & 31.87 & 73.62 & 10.38 & 16.44 & 76.63 & 16.56 & 36.67 & 76.87 & 8.50 &27.28 & 79.18   \\
Llama-3~\cite{llama3modelcard} & 11.79 & 34.82 & 77.37 & 13.82 & 18.95 & 85.80 & 18.78 & 40.20 & 78.73 & 12.25 & 35.16 & 82.79 \\
GPT-4~\cite{gpt4}         &          14.07 &           42.22 &            86.80 &           17.23 &           22.96 &           86.94 &           25.82 &           51.89 &           85.46 &           15.11 &           41.39 &           83.99 \\ 
GPT-4o~\cite{gpt4o} & 15.11 & 42.59 & 85.93 & 17.20 &  22.82 & 85.31 & 27.28 & 52.57 & 85.90 & 16.28 & 42.40 & 83.82 \\
\hline

\textit{\textbf{Prior Ensemble Method}}\\ \hdashline[3pt/3pt]
LLM-Blender~\cite{llm-blender} & 8.77 & 28.74 & 82.80 & 0.03 & 0.85 & 42.77 & 11.80 & 29.85 & 67.95 & 0.94 & 2.69 & 46.49 \\
EVA~\cite{eva} & 2.53 & 15.26 & 39.00 &1.51 & 3.57 &  37.17 & 9.77 &  28.40 &  68.75 & 7.99 & 27.00 & 73.15 \\
MBR~\cite{mbr} & 14.10 & 42.24 & 86.70 & 17.14 & 23.00 & 87.53 & 25.45 & 51.78 & 85.55 & 14.66 & 41.11 & 83.93 \\ \hline

\textit{\textbf{Proposed Method}}\\ \hdashline[3pt/3pt]
\ours (Llama-3; top3) & 15.60 & 39.86 & 84.10 & 14.56 & 19.92 & 87.34 & 23.41 & 45.95 & 81.66 & 14.27 & 38.25 & 81.80 \\
\ours (Llama-3; \textit{D, E}) & 13.85 & 37.36 & 69.96 & 14.97 & 20.21 & 85.42 & 21.35 & 43.75 & 79.71 & 12.37 & 36.51 & 82.09 \\

\ours (GPT-4; top3)   &          16.66 &           42.85 &  \textbf{86.82} &           17.95 &           23.84 &            87.50 &           27.22 &           51.73 &  85.65 &           16.53 &           42.41 &           84.46 \\

\ours (GPT-4; \textit{D, E})     &  17.10 &  43.29 &           85.92 &  18.18 &  24.05 &  \textbf{88.74} &  27.98 &  52.41 &           85.27 &  17.02 &  43.02 &  \textbf{84.82} \\

\ours (GPT-4o; top3) &
17.77 & 43.38 & 85.46 &
18.08 & 23.98 & 88.15 &
28.62 & 52.53 & 85.87 &
16.92 & 42.93 & 84.52 \\
 
\ours (GPT-4o; \textit{D, E}) &
\textbf{18.02} & \textbf{43.46} & 86.19 &
\textbf{18.31} & \textbf{24.32} & 88.33 &
\textbf{29.50} & \textbf{53.16} & \textbf{86.03} &
\textbf{17.66} & \textbf{43.73} & 84.27 \\

\Xhline{3\arrayrulewidth}
\end{tabular}
\end{adjustbox}
\caption{Main results. The best scores in each pair are marked \textbf{bold}. Within parentheses in the proposed method, the parts separated by semicolons denote the merging module and the candidates used. \textit{D} and \textit{E} represent candidates obtained from direct translation and English pivot, respectively.} 
\label{tab: main results}
\end{table*}

\subsection{Evaluation Metrics}
We assess the translation quality using BLEU~\cite{papineni-etal-2002-bleu}, chrF++~\cite{popovic-2017-chrf}, and reference-based COMET (\textit{wmt22-COMET-da}) \cite{comet22}.
For reporting BLEU, \textit{SacreBLEU}~\cite{post-2018-call} is used with the \texttt{ko-mecab} tokenizer for Korean and \texttt{13a} tokenizer for the others.

\subsection{Baselines}
\label{sec:baselines}

As an encoder-decoder NMT model, we use NLLB-200-distilled-600M~\cite{nllb}.
When training NLLB, we use the Transformers library from HuggingFace~\cite{wolf2020huggingfaces}.
AdamW optimizer~\cite{adamw} is used with a learning rate of $2e$$-$$5$, batch size of 2, and dropout with a probability of 0.1.
When validation BLEU did not improve for 3 checkpoints, with 30k steps between them, we stopped training.

For open-source LLMs, we use Vicuna 13B~\cite{Vicuna}, Baize 13B~\cite{baize}, and Llama-3-8B-Instruct~\cite{llama3modelcard} as baselines. 
We fine-tuned these LLMs with QLoRA~\cite{dettmers2024qlora}; $r$=16, $\alpha$=64, dropout=0.1 for all linear layers.
For black-box LLMs, we use GPT-4~\cite{gpt4} and GPT-4o~\cite{gpt4o}.
The versions \texttt{gpt-4-1106-preview} and \texttt{gpt-4o-2024-08-06} are employed for GPT-4 and GPT-4o, respectively.
For GPT models, \textit{temperature} is set to 0.0 for stable responses~\cite{peng2023making} and \textit{top\_ p} is set to 0.1 to ensure reproducibility. 
For LLMs, we use the zero-shot prompt template of~\citet{howgood}, as presented in Appendix~\ref{sec:apdx_prompt_templates}.

As state-of-the-art ensemble baselines, we employ \blender~\cite{llm-blender}, EVA~\cite{eva}, and MBR~\cite{mbr}.
For \blender and EVA, we fine-tuned the same open-source LLMs used in each study with the train data in Table~\ref{tab:dataset statistics}.
The list of the LLMs is in Appendix~\ref{sec:apdx_llms}.
\textit{temperature} is set to 0.1 to mitigate hallucination for low-resource pairs~\cite{guerreiro2023hallucinations}.
For MBR, we generate a set of 5 hypotheses using GPT-4.
When generating hypotheses, \textit{temperature} was set to 0.0 for its optimal performance, based on the results of our pilot experiments in Appendix~\ref{sec:MBR} and prior study~\cite{peng2023making}.
Other configurations are the same as in the original work~\cite{mbr}.

\subsection{Implementation Details}

In the candidate generation step of \ours, we employ \nllb.
For each source-target language pair, we use an NLLB fine-tuned for the language pair in Table~\ref{tab:dataset statistics} to generate the directly translated candidates.
For the merging module, we use Llama-3, GPT-4, and GPT-4o.
For all models used in \ours, including NLLB, Llama-3, GPT-4, and GPT-4o, we apply the same settings in \S\ref{sec:baselines}.

As detailed in \S\ref{sec:candidate aggregation}, we explore two approaches in the ensemble process: one dynamically selects the top-$\textit{k}$ ($\textit{k}$=3) candidates, and another uses candidates obtained from fixed paths.
To select the top-$\textit{k}$ candidates for each source sentence, we use the reference-free COMETkiwi.
When selecting candidates from fixed paths, we use directly translated candidates and English-pivoted candidates, which were the top-performing paths on the FLORES-200 benchmark.

\subsection{Main Results}

\begin{table}[t]
\centering
\scriptsize
\renewcommand{\arraystretch}{1.1}
\begin{adjustbox}{width=\columnwidth, center}
\begin{tabular}{lccc}
\Xhline{3\arrayrulewidth}
\multirow{2}{*}{\textbf{Model}}  & \multicolumn{3}{c}{\textbf{Korean$\rightarrow$Italian}}\\ \cline{2-4}
 & BLEU & chrF++ & COMET \\ \hline\hline 
\textit{\textbf{Candidate}}\\\hdashline[3pt/3pt]
 \nllb (direct)   & 16.27 & 41.14 & 84.60 \\
 \nllb (Portuguese pivot) & 13.13 & 37.57 & 83.21  \\  
 \nllb (Spanish pivot) & 13.87 & 38.47 & 83.71 \\
 \nllb (English pivot) & 14.77 & 39.39 & 81.48 \\
\Xhline{3\arrayrulewidth}
\end{tabular}
\end{adjustbox}
\caption{Quality of ensemble candidates.}
\label{tab:candidates}
\end{table}

Table~\ref{tab: main results} reports the overall performance of \ours and other methods.
The results show that \ours consistently outperforms baselines across all language pairs.
While standalone NMT systems rely solely on pre-trained knowledge, \ours explicitly leverages candidates during ensemble.
Even when training an open-source LLM, Llama-3, translation quality improves by utilizing candidates obtained via pivoting.
Compared to LLM-based translation, performance improves at minimal cost with a small 0.6B model.
Table~\ref{tab:candidates} presents the quality of candidates used in the ensemble.

\minisection{Comparison with multi-model ensemble}
We compare \ours with \blender~\cite{llm-blender} and EVA~\cite{eva}, state-of-the-art ensemble methods utilizing multiple models.
\blender employs $N$ ($N$=11) LLMs for candidate generation, picks top-3 candidates with \textsc{PairRanker}, and fuses them with \textsc{GenFuser}.
EVA is a token-level ensemble method that leverages vocabulary alignment across multiple models.

Results in Table~\ref{tab: main results} show that \ours outperforms multi-model ensemble baselines by a considerable margin.
\blender was unable to improve outputs compared to its candidate LLMs in non-English translation tasks.
Additionally, LLMs used for generating candidates in \blender, such as Vicuna and Baize, exhibit subpar performance on given tasks.
These results align with recent work~\cite{alma}; open-source LLMs often struggle when not translating into English.

EVA is not only ineffective on the given tasks but also has several limitations inherent to its design as a token-level ensemble.
First, EVA is unable to use black-box models such as GPT-4.
Second, it is memory-intensive, as it requires loading multiple models into memory simultaneously.
While multi-model ensemble methods generate candidates using up to 11 LLMs (with sizes up to 13B), \ours generates candidates with a significantly smaller single model (0.6B), thereby greatly reducing computational overhead.

\begin{table}[t!]
\centering
\small
\renewcommand{\arraystretch}{1.1}
\begin{adjustbox}{width=\columnwidth, center}
\begin{tabular}{lcccccc}
\Xhline{3\arrayrulewidth}
\textbf{Model} & BLEU & chrF++ & COMET & BLEU & chrF++ & COMET \\ \hline\hline 
& \multicolumn{6}{c}{\textbf{Distant Language Pairs}} \\ \cline{2-7}
& \multicolumn{3}{c}{Portuguese$\rightarrow$Russian} & \multicolumn{3}{c}{Russian$\rightarrow$Portuguese} \\
NLLB&	25.17&	51.77&	90.12&	29.69&	55.81&	86.01 \\
GPT-4&	26.50&	52.76&	91.11&	25.51&	54.05&	86.69 \\
\ours&	\textbf{27.48}&	\textbf{53.49}&	\textbf{91.74}&	\textbf{30.82}&	\textbf{56.73}&	\textbf{88.37} \\ \hline
& \multicolumn{3}{c}{Dutch$\rightarrow$Russian} & \multicolumn{3}{c}{Russian$\rightarrow$Dutch} \\
NLLB&	22.95&	50.21&	89.92&	25.56&	53.60&	88.18 \\
GPT-4&	24.37&	51.32&	91.28&	24.46&	53.85&	88.58 \\
\ours&	\textbf{25.45}&	\textbf{52.16}&	\textbf{91.47}&	\textbf{28.05}&	\textbf{55.80}&	\textbf{89.35} \\ \hline
& \multicolumn{3}{c}{French$\rightarrow$Ukrainian} & \multicolumn{3}{c}{Ukrainian$\rightarrow$French} \\
NLLB&	14.58&	37.11&	82.99&	20.69&	44.04&	80.61 \\
GPT-4&	13.84&	39.03&	84.12&	23.30&	47.13&	83.43 \\
\ours&	\textbf{17.20}&	\textbf{39.82}&	\textbf{86.55}&	\textbf{24.35}&	\textbf{47.17}&\textbf{84.36} \\ \hline \hline
& \multicolumn{6}{c}{\textbf{Similar Language Pair (Romance)}} \\ \cline{2-7}
& \multicolumn{3}{c}{Spanish$\rightarrow$Portuguese} & \multicolumn{3}{c}{Portuguese$\rightarrow$Spanish} \\
NLLB&	32.38&	56.97&	86.88&	33.63&	57.61&	85.13 \\
GPT-4&	29.94&	55.26&	84.84&	34.70&	58.63&	86.75 \\
\ours&	\textbf{34.06}&	\textbf{58.11}&	\textbf{87.70}&	\textbf{36.03}&	\textbf{59.32}&	\textbf{86.92} \\ \hline
& \multicolumn{6}{c}{\textbf{Similar Language Pair (Slavic)}} \\ \cline{2-7}
& \multicolumn{3}{c}{Ukrainian$\rightarrow$Russian} & \multicolumn{3}{c}{Russian$\rightarrow$Ukrainian} \\
NLLB&	22.16&	45.41&	89.82&	19.67&	43.35&	89.87 \\
GPT-4&	24.41&	\textbf{47.59}&	89.43&	\textbf{22.42}&	\textbf{45.61}&	90.39 \\
\ours&	\textbf{24.64}&	47.51&	\textbf{90.78}&	22.09&	45.40&	\textbf{90.70} \\ 
\Xhline{3\arrayrulewidth}
\end{tabular}
\end{adjustbox}
\caption{Results on all language pairs.}
\label{tab:Results on all translation directions}
\end{table}

\minisection{Results on all language pairs}
To validate generalizability, we report the results for all language pairs we experimented with, including those within the same language family.
Distant pairs refer to languages that belong to different families and use different scripts, while similar pairs belong to the same family and share the same script.
The statistics for each language pair are in Appendix~\ref{sec:apdx_dataset statistics}.
Language pairs used in the experiments are as follows:

\begin{itemize}
    \item Distant language pairs: Portuguese$\leftrightarrow$Russian, Dutch$\leftrightarrow$Russian, and French$\leftrightarrow$Ukrainian

    \item Similar language pairs: Spanish$\leftrightarrow$Portuguese and Ukrainian$\leftrightarrow$Russian
\end{itemize}

Table~\ref{tab:Results on all translation directions} shows the results with the top-performing baselines, NLLB~\cite{nllb} and GPT-4~\cite{gpt4}.
\ours consistently exhibits superior performance compared to strong baselines on distant language pairs. 
Surprisingly, it also showed improvements in similar language pairs, such as Spanish$\leftrightarrow$Portuguese.

\begin{table*}[t]
\centering
\large
\renewcommand{\arraystretch}{1.2}
\begin{adjustbox}{width=\textwidth, center}
\begin{tabular}{llc}
% \begin{tabular}{|p{1cm}|p{2cm}|p{5cm}|}
\Xhline{3\arrayrulewidth}
\textbf{\#} & \textbf{Type}& \textbf{Example}\\ \hline\hline

\multirow{8}{*}{\textbf{1}} & Source Sentence & 그래서 그동안 자문해왔습니다. 왜 우리는 질병들과 싸우기에 더 현명하고, 정확하며 더욱 적합한 ... \\ % 더욱 적합한 이 방법을 선진국에만 국한해야만 하는가? \\
& & \textit{(English Translation: So we've been asking ourselves, why should we limit this smarter, more precise, more appropriate ...)} \\ % , more appropriate approach to fighting diseases to the developed world?)}\\
& Target Reference & Quindi \hlc[lightblue]{mi sono chiesta}: perché dovremmo limitare questo modo intelligente, preciso, migliore ... \\ \cline{2-3} % , migliore di combattere le malattie ai paesi ricchi? \\ \cline{2-3}
& Top-1 Candidate & Quindi \hlc[lightblue]{ci siamo chiesti}, perché dovremmo limitare questo modo più intelligente, più preciso e più appropriato ... \\ % e più appropriato di combattere le malattie ai paesi sviluppati? \\
& Top-2 Candidate & Quindi \hlc[lightblue]{ci siamo chiesti}: perché dovremmo limitare questo metodo più intelligente, più preciso e più adatto ... \\ % e più adatto per combattere la malattia ai paesi sviluppati? \\ 
& Top-3 Candidate & Quindi nel corso di questo tempo, \hlc[pink]{abbiamo chiesto}: perché dovremmo limitare questo metodo più intelligente, più preciso e più adeguato ... \\ \cline{2-3}% e più adeguato per combattere le malattie ai paesi sviluppati? \\ \cline{2-3}
& GPT-4 & Quindi, \hlc[pink]{abbiamo cercato consigli} fino ad ora. Perché dobbiamo limitare questo metodo, che è più intelligente, preciso e più adatto ... \\ % e più adatto per combattere le malattie, solo ai paesi sviluppati? \\
& \ours  & Quindi \hlc[lightblue]{ci siamo chiesti}: perché dovremmo limitare questo metodo più intelligente, più preciso e più adatto ... \\ \hline\hline % e più adatto per combattere le malattie ai paesi sviluppati? \\ \hline\hline

\multirow{8}{*}{\textbf{2}} & Source Sentence & 많은 사람들이 헤드폰을 사용하는데 이것의 문제점은 3가지 큰 건강 이슈를 가져온다는 것입니다. \\
& & \textit{(English Translation: The trouble with widespread headphone use is it brings three really big health issues.)}\\
& Target Reference & Il problema dell'utilizzo diffuso degli auricolari è che scatenano tre grandi \hlc[lightblue]{problemi} di salute. \\ \cline{2-3}
& Top-1 Candidate & Il problema è che molte persone usano le cuffie, e questo Porta a tre grandi \hlc[lightblue]{problemi} di salute. \\
& Top-2 Candidate & Il problema è che molte persone usano le cuffie, e questo Porta a tre grandi \hlc[lightblue]{problemi} di salute. \\ 
& Top-3 Candidate & Il problema è che molte persone usano cuffie, e questo è ciò che causa tre \hlc[lightblue]{problemi} di salute principali. \\ \cline{2-3}
& GPT-4 & Molte persone utilizzano le cuffie, ma il problema è che ciò comporta tre importanti \hlc[pink]{questioni} di salute. \\
& \ours  &  Il problema è che molte persone usano le cuffie, e questo porta a tre grandi \hlc[lightblue]{problemi} di salute. \\

\Xhline{3\arrayrulewidth} 
\end{tabular}
\end{adjustbox}
\caption{Case study. Parts equivalent to the source sentence and mistranslated parts are highlighted in \hlc[lightblue]{blue} and \hlc[pink]{red}, respectively. English translation of the source comes from another pair within the dataset.}
\label{tab:case study}
\end{table*}

\minisection{Case study}
We conduct a qualitative analysis to verify the impact of candidates on the final translation. 
We compare the output of GPT-4, used as the merging module, with \ours, which utilizes candidates for the ensemble process.
In Table~\ref{tab:case study}, we provide two examples along with the source and target sentences, as well as the top-3 candidates.

In the first example, we observe that \ours appropriately translates homonyms in context.
In Korean, ``자문'' has the meaning of both ``consultation'' and ``asking oneself''.
Considering the context, the expression should convey the meaning of ``asking ourselves'', as also shown in the English translation.
However, GPT-4 mistranslated the source sentence, converting the phrase ``자문해왔습니다'' to ``abbiamo cercato consigli'' (``seeking consultation from others'').
On the other hand, \ours accurately translates ``ci sono chiesti'', meaning ``asking ourselves'', aligning well with the context by leveraging information from candidates.

In the second example, GPT-4 translates the source sentence by translating the noun ``이슈'' into ``questioni''.
However, given the topic of discussing potential health risks, this translation does not fit well with the overall context.
By contrast, the ensemble result of \ours, generated using the identical model, improves translation quality by using a more accurate expression ``problemi'', despite having access to the same pre-trained knowledge.
Additionally, when more suitable expressions (e.g., ``ne vale la pena'') appear in candidates, \ours utilizes them to refine the final translation.

\subsection{Analysis}

\begin{table}[t]
\centering
\large
\begin{adjustbox}{width=\columnwidth, center}
\renewcommand{\arraystretch}{1.06}
\begin{tabular}{lcccc}
\Xhline{3\arrayrulewidth}
\multirow{2}{*}{\parbox{4cm}{\textbf{Cand. Generation}}} & \multirow{2}{*}{\parbox{1.5cm}{\textbf{\# Cand.}}}  &  \multicolumn{3}{c}{\textbf{Korean$\rightarrow$Italian}} \\ \cline{3-5} 
 & & BLEU & chrF++ & COMET \\ \hline\hline 
LLMs (\blender) & 11 &  14.75 & 41.29 & 86.20\\ 
LLMs + \nllb (direct) & 12 & 16.08 & 42.38 & 86.22 \\ 
\nllb (pivot, ours) & 4 & \textbf{16.66} & \textbf{42.85} & \textbf{86.82}  \\
\Xhline{3\arrayrulewidth}
\end{tabular}
\end{adjustbox}
\caption{Candidate generation method comparison.}
\label{tab:comparison changing candidates}
\end{table}

\begin{table}[!t]
\centering
\renewcommand{\arraystretch}{1.06}
\begin{adjustbox}{width=\columnwidth, center}
\begin{tabular}{lccc}
\Xhline{3\arrayrulewidth}
\multirow{2}{*}{\textbf{Model}}  & \multicolumn{3}{c}{\textbf{Korean$\rightarrow$Italian}}\\ \cline{2-4}
 & BLEU & chrF++ & COMET \\ \hline\hline 
\textit{\textbf{Standalone NMT System}}\\\hdashline[3pt/3pt]
\nllb~\cite{nllb}   & 16.27 & 41.14 & 84.60 \\ \hline
\textit{\textbf{Encoder-Decoder}}\\\hdashline[3pt/3pt]
{\ours} (FiD)      & 13.74 & 36.78 & 78.98 \\
{\ours} (TRICE)    & 15.89 & 41.98 & 84.06 \\\hline
\textit{\textbf{LLM-based}}\\\hdashline[3pt/3pt]
{\ours} (\textsc{GenFuser}) & 14.56 & 39.32 & 80.07\\
{\ours} (GPT-4)    & \textbf{16.66} & \textbf{42.85} & \textbf{86.82}\\
\Xhline{3\arrayrulewidth}
\end{tabular}
\end{adjustbox}
\caption{Evaluation of merging module variants.}
\label{tab:merging module results}
\end{table}

\begin{figure*}[t!]
  \centering
  \begin{subfigure}[b]{0.76\textwidth}
    \centering
    \includegraphics[width=\textwidth,height=3.7cm]{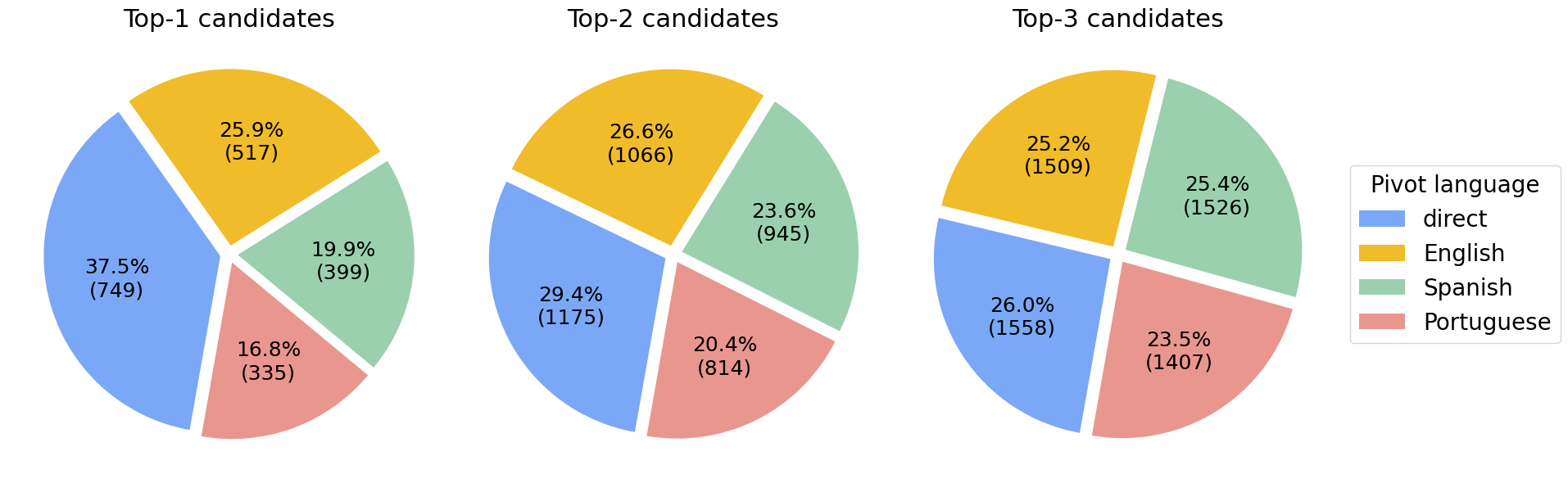}
    \caption{Proportion of pivot languages (Korean$\rightarrow$Italian) comprising the top-$\textit{k}$ candidates.}
    \label{fig:top-k candidates}
  \end{subfigure}
  \hfill
  \begin{subfigure}[b]{0.22\textwidth}
    \centering
    \includegraphics[width=\textwidth,height=3.7cm]{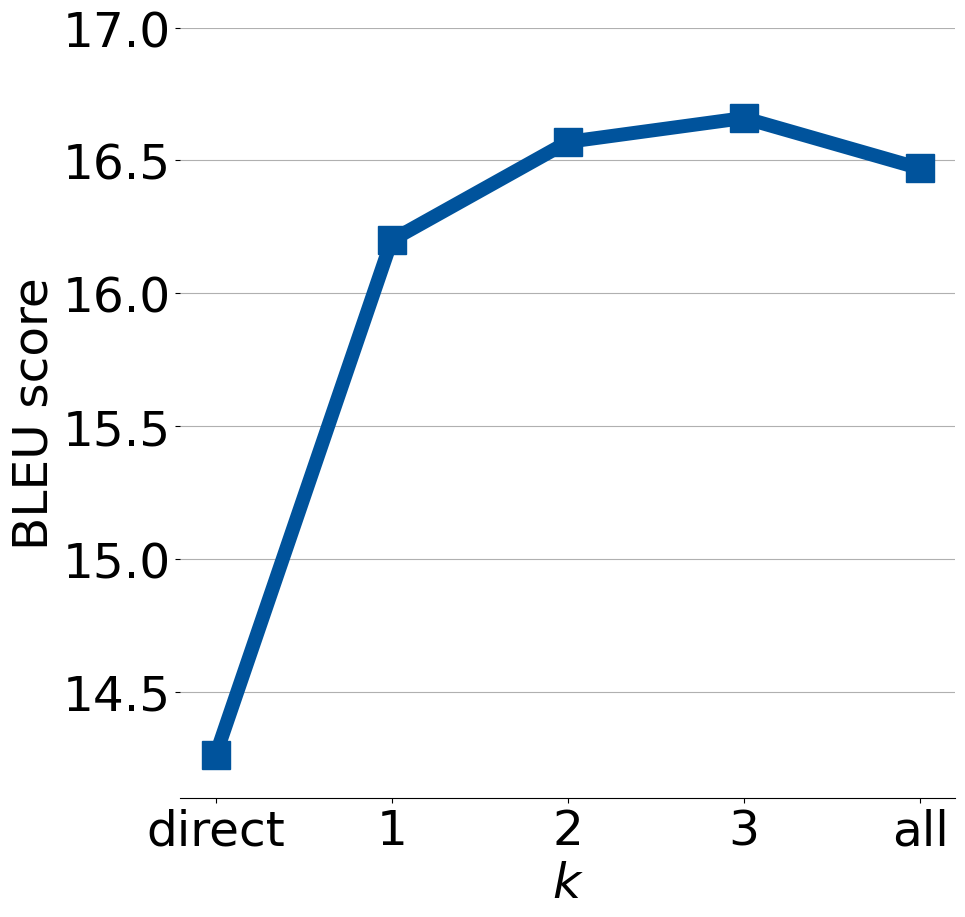}
    \caption{Impact of top-$\textit{k}$ values on performance.}
    \label{fig:top-k bleu score}
  \end{subfigure}
  \caption{Analysis of pivot language distribution and top-$k$ performance.}
  \label{fig:top-k-analysis}
\end{figure*}

\minisection{Candidate generation}
To validate the effectiveness of \ours, we conduct experiments only varying the candidate generation method, while using the same merging module.
We compare a candidate pool of size 4 obtained through pivot translation (\ours) with a candidate pool of size 11 obtained using 11 LLMs in \blender.

As shown in Table~\ref{tab:comparison changing candidates}, the proposed method of generating candidates through pivot translation achieves the highest performance, despite using the smallest candidate pool.
From the perspective of direct translation in NLLB, leveraging 3 candidates obtained through pivot translation yields higher scores than incorporating candidates generated by 11 LLMs.
These results demonstrate that using stable-quality candidates generated by a single model via pivot translation outperforms the use of multiple models with performance disparities.

\minisection{Candidate aggregation}
We first investigate whether \ours shows improvement when utilizing other merging modules.
As detailed in \S\ref{sec:candidate aggregation}, we run experiments with three architectures: FiD~\cite{fid}, TRICE~\cite{trice}, and \textsc{GenFuser}~\cite{llm-blender}.
When implementing FiD, we replace the backbone of FiD with $\text{mT5}_{\texttt{BASE}}$~\cite{xue-etal-2021-mt5}.
TRICE is a method proposed for multi-source translation. 
Since TRICE was not originally intended for ensemble use, we repurposed it by training on the following two tasks:
The first task is the original translation, which converts source sentences into target sentences.
The second task is refining candidates that are paired with target references.
In the case of \trice, only the highest quality candidates, which are the directly translated ones, are used due to its architecture.
\fid and \textsc{GenFuser} use top-3 candidates.

Table~\ref{tab:merging module results} shows that the ensemble methods using encoder-decoder architectures and \textsc{GenFuser} do not yield improved results.
These methods struggle to leverage additional information from the candidates and, consequently, do not enhance performance. 
In contrast, using GPT-4 as the merging module leads to better performance compared to the standalone NMT system.

We also compare ranking methods COMETkiwi~\cite{rei2022cometkiwi} and \textsc{PairRanker}~\cite{llm-blender}.
Table~\ref{tab:candidates selecting method} compares the results after selecting the top-3 using \textsc{PairRanker} and COMETkiwi.
As shown in the results, the difference is not significant.
We believe this is because the candidates selected by both ranking methods are similar.
There are 979 out of 2000 test sentences (48.95\%) where the top-3 candidates selected by both ranking methods are the same. 
In cases with 2 out of 3 matches, there are 1533 instances (76.65\%).
Given the similarity in predictions by both ranking methods, the final scores exhibit comparable performance, except in the case of COMET.
From the cost perspective, \textsc{PairRanker} requires comparisons for $O(N^2)$ unique pair combinations depending on the number of candidates $N$.
However, COMETkiwi only needs to sort the scores of $N$ candidates, resulting in a time complexity of $O(N\log N)$.
Therefore, due to its computational efficiency, we use COMETkiwi to rank candidates.

\begin{table}[t!]
\centering
\small
\begin{adjustbox}{width=\columnwidth, center}
\renewcommand{\arraystretch}{1.1}
\begin{tabular}{lccc}
\Xhline{3\arrayrulewidth}
\multirow{2}{*}{\textbf{Method}} &  \multicolumn{3}{c}{\textbf{Korean$\rightarrow$Italian}} \\ \cline{2-4} 
 & BLEU & chrF++ & COMET \\ \hline\hline 
\textsc{PairRanker}~\cite{llm-blender} & \textbf{16.74} & 42.82 & 85.92\\
COMETkiwi~\cite{rei2022cometkiwi} & 16.66 & \textbf{42.85} & \textbf{86.82} \\
\Xhline{3\arrayrulewidth}
\end{tabular}
\end{adjustbox}
\caption{Impact of candidate ranking strategies.}
\label{tab:candidates selecting method}
\end{table}

\minisection{Comparison with selection-based ensemble}
With a selection-based ensemble, we can choose one of the existing candidates as the final translation, rather than generating a new one.
In this experiment, we compare our approach with a selection-based ensemble by selecting the top-1 translation using \textsc{PairRanker}~\cite{llm-blender} and COMETkiwi~\cite{rei2022cometkiwi}.
Additionally, we report results with an ideal case: selecting top-1 by considering references as well, which are not available in practice.
The ideal top-1 is selected by reference-based COMET~\cite{comet22}.

As shown in Table~\ref{tab:selection-based}, \ours exhibits superior performance compared to the selection-based ensemble methods.
Even when we leverage reference-based COMET, which is impossible in real-world scenarios due to the necessity for references, \ours outperforms it in chrF++ and COMET.
These results indicate that performing a generation-based ensemble with pivoting can effectively produce final translations that surpass those selected from the existing candidate pool.

\begin{table}[t]
\centering
\small
\begin{adjustbox}{width=\columnwidth, center}
\renewcommand{\arraystretch}{1.1}
\begin{tabular}{llccc}
\Xhline{3\arrayrulewidth}
\multirow{2}{*}{\textbf{Category}} & \multirow{2}{*}{\textbf{Method}} &  \multicolumn{3}{c}{\textbf{Korean$\rightarrow$Italian}} \\ \cline{3-5} 
& & BLEU & chrF++ & COMET \\ \hline\hline 
\multirow{3}{*}{\makecell[l]{Selection-based \\ (top-1)}} & \textsc{PairRanker} & 15.61 & 40.62 & 84.46 \\
& COMETkiwi & 15.61 & 40.71 & 84.10 \\
& COMET* (ideal) & \textbf{17.77} & 42.81 & 84.83\\ \hline
Generation-based & \ours & \underline{16.66} & \textbf{\underline{42.85}} & \textbf{\underline{86.82}} \\
\Xhline{3\arrayrulewidth}
\end{tabular}
\end{adjustbox}
\caption{Compare with selection-based ensemble. COMET* is the ideal baseline, as it requires references. Best scores including COMET* are \textbf{bolded}, while best scores excluding it are \underline{underlined}.}
\label{tab:selection-based}
\end{table}

\minisection{Analysis of selected candidates}
We conduct experiments to investigate the impact of the value of $\textit{k}$ in the top-$\textit{k}$ candidates and their composition.
Figure~\ref{fig:top-k candidates} illustrates the proportion of pivot languages composing the top-$\textit{k}$ candidates.
Top-$\textit{k}$ candidates, selected by the QE metric, are composed of diverse candidates obtained through various pivot languages.
We also observe the same tendency in other datasets.
This suggests that generating diverse candidates through multiple paths helps acquire higher-quality candidates.

Figure~\ref{fig:top-k bleu score} presents BLEU for different values of the $\textit{k}$.
The highest BLEU is achieved when $k$ is set to 3.
These results demonstrate that more candidates in the aggregation process increase the likelihood of providing contextually appropriate information.
However, it converges around top-3, which we attribute to the inclusion of lower-scoring candidates such as degenerated sentences.
Hence, as $k$ increases, the improvement reaches a plateau.

\section{Conclusion}

In this work, we introduced \ours, a pivot-based single model ensemble framework, to enhance low-resource language translation.
By transferring knowledge from diverse pivot languages, we were able to obtain not only diverse but also high-quality candidates.
Since the optimal path to generating the best candidate varies per sentence, our study underscores the significance of exploiting a spectrum of pivot languages.
Moreover, the single model generation process offers cost savings compared to multi-model ensemble approaches. 
Empirical results and qualitative analyses show that the proposed method can yield contextually suitable translations for the given source sentences by leveraging pivoted candidates.
\section{Limitations}

Although \ours utilizes candidates obtained via pivoting, limitations arise from the nature of pivot translation.
Constraining the pivot languages to high-resource languages can limit the number of candidates because pivoting through low-resource languages can lead to some loss of information due to error propagation inherent in the two-step translation.
This semantic shift potentially causes a decrease in candidate quality.
If the quality of candidates declines, improvements from the ensemble might not be significant, indicating a limitation in the number of pivot paths.
\section{Acknowledgements}

This work was supported by the National Research Foundation of Korea (NRF) grant funded by the Korea government (MSIT) (RS-2025-16070246).
This work was supported by the National R\&D Program for Cancer Control through the National Cancer Center(NCC) funded by the Ministry of Health \& Welfare, Republic of Korea (No.RS-2025-02264000).

\section{Bibliographical References}\label{sec:reference}

\bibliographystyle{lrec2026-natbib}
\bibliography{lrec2026-pivote}

\section{Language Resource References}
\label{lr:ref}
% \bibliographystylelanguageresource{lrec2026-natbib}
% \bibliographylanguageresource{languageresource}

\section{Appendices}
\appendix

\section{Pivot Language Selection}
\label{sec:apdx_top4 pivot langauges}

Based on the results from the FLORES-200~\cite{nllb} benchmark, we select the top-4 pivot paths as presented in Table~\ref{tab:pivot path}.
We utilize the full 2009 sentences as our test set: 997 sentences from the \textit{dev} and 1012 sentences from the \textit{devtest}.
The pivot language pool is chosen as the \textit{bridge languages} in \citet{m2m100}.

\begin{table}[h]
\centering
\renewcommand{\arraystretch}{1.1}
\begin{adjustbox}{width=\columnwidth, center}
\begin{tabular}{ccccc}
\Xhline{3\arrayrulewidth}

\multirow{2}{*}{\textbf{Pivot Language}} & \multicolumn{4}{c}{\textbf{Lang-pair}} \\ \cline{2-5} 
 & KO$\rightarrow$IT& IT$\rightarrow$KO & AR$\rightarrow$PT & PT$\rightarrow$AR \\ \hline\hline 

\texttt{direct}    & \textbf{14.02}& \textbf{18.63} & \textbf{27.15} & \textbf{15.22} \\
\texttt{arb\_Arab} & 11.03 & 15.82 & - & - \\
\texttt{ben\_Beng} & 10.79 & 15.44 & 18.65 & 9.76 \\
\texttt{ces\_Latn} & 11.48 & 16.08 & 21.23 & 11.55 \\
\texttt{deu\_Latn} & 12.49 & 17.11 & 22.62 & 12.56 \\
\texttt{ell\_Grek} & 11.96 & 16.54 & 22.53 & 12.54 \\
\texttt{eng\_Latn} & \textbf{14.82} & \textbf{19.34} & \textbf{28.40} & \textbf{15.92} \\
\texttt{fin\_Latn} & 9.62 & 14.31 & 17.27 & 9.48 \\
\texttt{fra\_Latn} & 13.55 & 17.27 & \textbf{24.96} & 13.77 \\
\texttt{heb\_Hebr} & 10.42 & 14.37 & 20.31 & 10.94 \\
\texttt{hin\_Deva} & 11.54 & 17.12 & 21.79 & 11.72 \\
\texttt{hun\_Latn} & 10.54 & 14.96 & 18.64 & 9.65 \\
\texttt{ind\_Latn} & 12.41 & 17.03 & 22.47 & 11.97 \\
\texttt{ita\_Latn} & - & - & 24.70 & \textbf{14.09} \\
\texttt{jpn\_Jpan} & 10.60 & 14.73 & 14.29 & 7.31 \\
\texttt{kor\_Hang} & - & - & 16.09 & 7.67 \\
\texttt{lit\_Latn} & 10.46 & 14.96 & 18.14 & 9.47 \\
\texttt{nld\_Latn} & 12.27 & 17.10 & 23.22 & 12.94 \\
\texttt{pes\_Arab} & 11.09 & 15.86 & 20.88 & 11.50 \\
\texttt{pol\_Latn} & 11.54 & 15.86 & 21.14 & 11.60 \\
\texttt{por\_Latn} & \textbf{13.80} & \textbf{18.01} & - & - \\
\texttt{rus\_Cyrl} & 12.25 & 16.57 & 22.77 & 12.39 \\
\texttt{spa\_Latn} & \textbf{13.89} & \textbf{18.39} & \textbf{26.60} & \textbf{14.91} \\
\texttt{swe\_Latn} & 11.93 & 16.54 & 22.34 & 12.25 \\
\texttt{swh\_Latn} & 10.66 & 14.22 & 19.13 & 10.19 \\
\texttt{tam\_Taml} & 9.90 & 14.92 & 18.09 & 9.48 \\
\texttt{tur\_Latn} & 11.25 & 15.92 & 19.53 & 10.04 \\
\texttt{ukr\_Cyrl} & 11.76 & 16.43 & 21.87 & 12.12 \\
\texttt{vie\_Latn} & 12.00 & 16.32 & 21.39 & 11.49 \\
\texttt{zho\_Hans} & 10.00 & 11.51 & 15.29 & 6.82 \\

\Xhline{3\arrayrulewidth} 
\end{tabular}
\end{adjustbox}
\caption{BLEU scores on FLORES-200 benchmark. Pivot languages are sorted in alphabetical order and top-4 pivot paths are marked \textbf{bold}.}
\label{tab:pivot path}
\end{table}

\begin{figure*}[!t]
  \centering
  \includegraphics[width=0.65\textwidth]{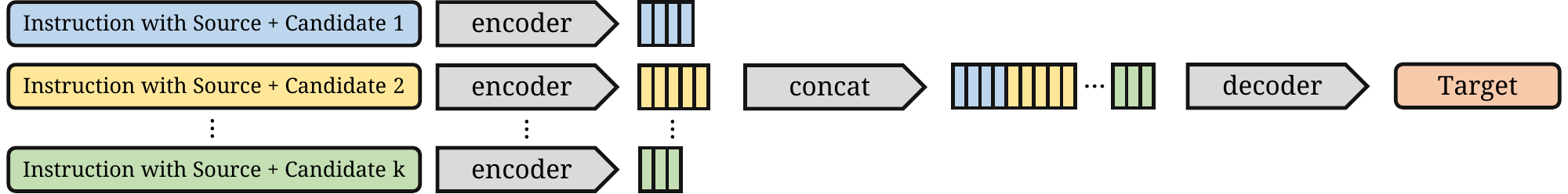} 
  \caption{Illustration of the merging process using FiD~\cite{fid}.}
  \label{fig:FiD}
\end{figure*}

\begin{figure*}[!t]
  \centering
  \includegraphics[width=0.5\textwidth]{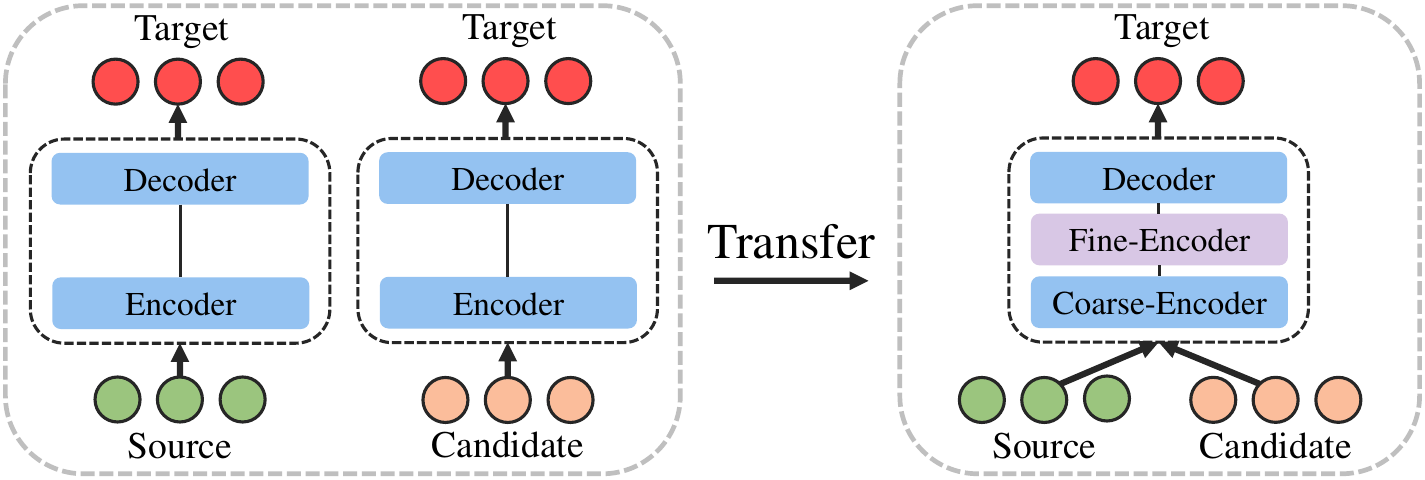} 
  \caption{Illustration of the merging process using TRICE~\cite{trice}.}
  \label{fig:TRICE}
\end{figure*}

\section{Impact of Resource-level of Pivot Languages}
\label{apdx:resource level of pivot languages}

Under the assumption that high-quality candidates are more adept at conveying the meaning of the source sentence, we select the top-4 paths based on scores on FLORES-200.
To verify this hypothesis, we conduct experiments using mid/low-resource pivot languages.
According to WMT22\footnote{\url{https://www.statmt.org/wmt22/translation-task.html}}, we select Ukrainian and Croatian as mid- and low-resource languages, respectively.
Table~\ref{tab:mid/low-resource pivot} shows that using candidates from high-resource languages outperforms those obtained from mid- and low-resource languages. 
The quality of candidates is presented in Table~\ref{tab:mid/low-resource candidates}.
In conclusion, since high-resource languages can also provide sufficient diversity, we select top-performing paths based on the results on FLORES-200.

\begin{table}[h]
\centering
\small
\begin{adjustbox}{width=\columnwidth, center}
\renewcommand{\arraystretch}{1.0}
\begin{tabular}{lccc}
\Xhline{3\arrayrulewidth}
\multirow{2}{*}{\textbf{Method}} &  \multicolumn{3}{c}{\textbf{Korean$\rightarrow$Italian}} \\ \cline{2-4} 
 & BLEU & chrF++ & COMET \\ \hline\hline 
\ours (GPT-4; \textit{U, C}) & 15.28 & 41.78 & 85.75 \\
\ours (GPT-4; \textit{E, S}) & \textbf{16.27} & \textbf{42.55} & \textbf{86.50} \\
\Xhline{3\arrayrulewidth}
\end{tabular}
\end{adjustbox}
\caption{Comparison with mid- and low-resource languages. \textit{U}, \textit{C}, \textit{E}, and \textit{S} represent candidates from Ukrainian, Croatian, English, and Spanish pivot, respectively.}
\label{tab:mid/low-resource pivot}
\end{table}

\begin{table}[h]
\centering
\small
\renewcommand{\arraystretch}{1.0}
\begin{adjustbox}{width=\columnwidth, center}
\begin{tabular}{lccc}
\Xhline{3\arrayrulewidth}

\multirow{2}{*}{\textbf{Model}}  & \multicolumn{3}{c}{\textbf{Korean$\rightarrow$Italian}}\\ \cline{2-4}
 & BLEU & chrF++ & COMET \\ \hline\hline 
                  
\textit{\textbf{Candidate}}\\\hdashline[3pt/3pt]
 \nllb (Ukrainian pivot)  & 11.95 & 35.03 & 82.32 \\  
 \nllb (Croatian pivot)   & 12.25 & 35.93 & 79.91 \\
 \nllb (Spanish pivot)    & 13.87 & 38.47 & 83.71 \\
 \nllb (English pivot)    & 14.77 & 39.39 & 81.48 \\

\Xhline{3\arrayrulewidth}
\end{tabular}
\end{adjustbox}
\caption{Quality of candidates from each pivot path.}
\label{tab:mid/low-resource candidates}
\end{table}

\section{Metric for Selecting Translation Paths}
\label{apdx: Metric for Selecting Translation Paths}

To analyze the impact of the path selection metric, we compare results using BLEU and COMET.
As Table~\ref{tab:ensemble results from BLEU/COMET} shows, the performance differences are marginal.
This is because both metrics select largely overlapping pivot languages, with only minor differences in ordering, as detailed in Table~\ref{tab:pivot languages from BLEU/COMET}.

\begin{table}[!h]
\centering
\small
\renewcommand{\arraystretch}{0.95}
\begin{adjustbox}{width=\columnwidth, center}
\begin{tabular}{lcccc}
\Xhline{3\arrayrulewidth}

\multirow{2}{*}{\textbf{top-$\textit{k}$}} & \multirow{2}{*}{\textbf{Path Selection}} & \multicolumn{3}{c}{\textbf{Korean$\rightarrow$Italian}}\\ \cline{3-5}
& & BLEU & chrF++ & COMET \\ \hline\hline 
\multirow{2}{*}{top-$\textit{1}$} & COMET	&15.98	&42.59	&86.22 \\
& BLEU	&16.20 &	42.84	&85.36 \\
\multirow{2}{*}{top-$\textit{2}$} & COMET	&16.46&	42.58&	86.66 \\
& BLEU	&16.57	&43.04&	86.37 \\
\multirow{2}{*}{top-$\textit{3}$} & COMET	&16.39&	42.41&	86.04 \\
& BLEU	&16.66	&42.85&	86.82 \\

\Xhline{3\arrayrulewidth}
\end{tabular}
\end{adjustbox}
\caption{Impact of the pivot path selection metric.}
\label{tab:ensemble results from BLEU/COMET}
\end{table}

\begin{table*}[!t]
\centering
\renewcommand{\arraystretch}{1.1}
\begin{adjustbox}{width=\textwidth, center}
\begin{tabular}{ccc}
\Xhline{3\arrayrulewidth}

\textbf{Lang-pair}  & \textbf{BLEU} & \textbf{COMET} \\ \hline\hline 
                  
KO$\rightarrow$IT & English (14.82), direct (14.02), Spanish (13.89), Portuguese (13.80)& English (82.89), Spanish (82.70), Indonesian (81.62), Portuguese (81.50) \\
IT$\rightarrow$KO & English (19.34), direct (18.63), Spanish (18.39), Portuguese (18.01)& Spanish (87.32), English (87.07), Portuguese (87.02), French (86.14) \\
AR$\rightarrow$PT & English (28.40), direct (27.15), Spanish (26.60), French (24.96)	 & direct (85.71), English (85.57), Spanish (85.54), Indonesian (84.94) \\
PT$\rightarrow$AR & English (15.92), direct (15.22), Spanish (14.91), Italian (14.09)	& French (82.65), direct (81.36), English (81.04), German (80.44) \\

\Xhline{3\arrayrulewidth}
\end{tabular}
\end{adjustbox}
\caption{Selected top-4 pivot paths from each metric. Scores are from experiments on FLORES-200.}
\label{tab:pivot languages from BLEU/COMET}
\end{table*}

\section{Prompt Templates}
\label{sec:apdx_prompt_templates}

We use the zero-shot prompt template from \citet{howgood} to instruct the LLMs for translation,

\begin{quote}

    \footnotesize
    \texttt{Translate this sentence from [source language] to [target language], Source: [source sentence]}

    \texttt{Target:}
\end{quote}

when ensembling with candidates, we use the prompt template as follows,

\begin{quote}
    \footnotesize
    \texttt{Ensemble the [source language] sentence with the provided [target language] candidates to create the best possible [target language] translation.}
    
    \texttt{[source language] sentence: [source sentence]}
    
    \texttt{[target language] candidate k: [target candidate]}
    
    \texttt{Please provide only the [target language] translation and no additional text.}
    
    \texttt{[target language] translation:}
\end{quote}

\section{Illustrations of the Merging Modules}
\label{apdx:Fid/TRICE illustration}

In this section, we provide visual illustrations of the encoder-decoder ensemble architectures, specifically FiD and TRICE. 
Figure~\ref{fig:FiD} and Figure~\ref{fig:TRICE} depict the input formatting and processing pipelines for each architecture, respectively.

\section{Open-source LLMs}
\label{sec:apdx_llms}
In experiments with \blender and EVA, we employ the same models as used in each paper.
These open-source LLMs are listed in Table~\ref{tab:11 open-source llms}.

\begin{table}[h!]
\centering
\small
\begin{adjustbox}{width=\columnwidth, center}
\renewcommand{\arraystretch}{1.03}
\begin{tabular}{lc}
\Xhline{3\arrayrulewidth}
\textbf{Model} & \textbf{Model Size} \\ \hline\hline
\textit{\textbf{\blender}~\cite{llm-blender}}\\ \hdashline[3pt/3pt]
Vicuna~\cite{Vicuna} & 13B \\
Baize~\cite{baize} & 13B\\
Alpaca~\cite{alpaca} & 13B \\ 
Koala~\cite{koala} & 13B \\
Open Assistant~\cite{oasst} & 12B \\
Dolly V2~\cite{dollyv2} & 12B \\
Flan-T5~\cite{Flan-T5} & 11B\\
MOSS~\cite{moss} & 7B \\
Mosaic MPT~\cite{MosaicML} & 7B \\ 
StableLM~\cite{stablelm} & 7B \\
ChatGLM~\cite{chatglm} & 6B \\ \hline

\textit{\textbf{EVA}~\cite{eva}}\\ \hdashline[3pt/3pt]
Baichuan2-Chat~\cite{baichuan2} & 7B \\
TigerBot-Chat-V3~\cite{tigerbot} & 7B \\
Vicuna-V1.5~\cite{Vicuna} & 7B \\
Llama-2-Chat~\cite{touvron2023llama2openfoundation} & 7B \\

\Xhline{3\arrayrulewidth}
\end{tabular}
\end{adjustbox}
\caption{Open-source LLMs along with their respective model sizes.}
\label{tab:11 open-source llms}
\end{table}

\section{Impact of \textit{temperature} in MBR}
\label{sec:MBR}

To investigate the best performance of MBR, we compared it across three different \textit{temperature} configurations: 1.0, 0.5, and 0.0, which were used in prior works by \citet{mbr}, \citet{temp0.5}, and \citet{peng2023making}, respectively.

Tables~\ref{tab:Impact of temperature in MBR decoding} and \ref{tab:Average quality of MBR hypotheses} show the quality of MBR outputs and hypotheses under different \textit{temperature} settings, respectively.
Aligning with the findings of the previous study~\cite{peng2023making}, we observed that a lower \textit{temperature} setting achieved better performance.
Thus, we set the \textit{temperature} to 0.0 for MBR in our experiments.

\begin{table}[h]
\centering
\renewcommand{\arraystretch}{1.2}
\begin{adjustbox}{width=\columnwidth, center}
\begin{tabular}{lccc}
\Xhline{3\arrayrulewidth}

\multirow{2}{*}{\textbf{Method}}  & \multicolumn{3}{c}{\textbf{Korean$\rightarrow$Italian}}\\ \cline{2-4}
 & BLEU & chrF++ & COMET \\ \hline\hline 
                  
 MBR (\textit{temp}=1.0;~\citet{mbr})   & 13.53 & 42.13 & 86.57 \\
 MBR (\textit{temp}=0.5;~\citet{temp0.5}) & 13.90 & 42.19 & 86.69 \\  
 MBR (\textit{temp}=0.0;~\citet{peng2023making})   & \textbf{14.10} & \textbf{42.24} & \textbf{86.70} \\ 

\Xhline{3\arrayrulewidth}
\end{tabular}
\end{adjustbox}
\caption{Impact of \textit{temperature} in MBR decoding.}
\label{tab:Impact of temperature in MBR decoding}
\end{table}

\begin{table}[h]
\centering
\small
\renewcommand{\arraystretch}{1.1}
\begin{adjustbox}{width=\columnwidth, center}
\begin{tabular}{lccc}
\Xhline{3\arrayrulewidth}

\multirow{2}{*}{\textbf{Method}}  & \multicolumn{3}{c}{\textbf{Korean$\rightarrow$Italian}}\\ \cline{2-4}
 & BLEU & chrF++ & COMET \\ \hline\hline 
                  
 \multirow{2}{*}{MBR hypotheses (\textit{temp}=1.0)}   & 13.47&	41.71 &84.94\\
  & (±0.21) &  (±0.14)	 & (±2.62) \\
 \multirow{2}{*}{MBR hypotheses (\textit{temp}=0.5)} & 13.86 &	 42.03&	86.55  \\  
  & (±0.13) &   (±0.11)& (±0.15)  \\
 \multirow{2}{*}{MBR hypotheses (\textit{temp}=0.0)} & 14.09 & 42.21 & 86.62 \\
  &   (±0.07) &  (±0.06)	   & (±0.10)   \\

\Xhline{3\arrayrulewidth}
\end{tabular}
\end{adjustbox}
\caption{Average quality of MBR hypotheses.}
\label{tab:Average quality of MBR hypotheses}
\end{table}

\section{Dataset Statistics}
\label{sec:apdx_dataset statistics}

Table~\ref{tab:apdx_dataset statistics} shows the dataset statistics for each language pair used in the experiments in Table~\ref{tab:Results on all translation directions}.

\begin{table}[h!]
\centering
\begin{adjustbox}{width=\columnwidth, center}
\renewcommand{\arraystretch}{1.2}
\begin{tabular}{ccccc}
\Xhline{3\arrayrulewidth}
\multirow{2}{*}{\textbf{Lang-pair}} & \multirow{2}{*}{\textbf{Dataset}} 
& \multicolumn{3}{c}{\textbf{\# Sentences}} \\ \cline{3-5} 
 & & \textbf{Train} & \textbf{Dev} & \textbf{Test} \\ \hline\hline 

\multicolumn{5}{c}{\textbf{Distant Language Pairs}} \\  \hline 

\multirow{2}{*}{PT $\leftrightarrow$ RU} & news-commentary v18.1 & \multirow{2}{*}{66,743} & \multirow{2}{*}{2,000} & \multirow{2}{*}{2,000} \\
 & \cite{opus} & & & \\ \hline  
 
 \multirow{2}{*}{NL $\leftrightarrow$ RU} & news-commentary v18.1 & \multirow{2}{*}{80,724} & \multirow{2}{*}{2,000} & \multirow{2}{*}{2,000} \\
 & \cite{opus} & & & \\ \hline

 \multirow{2}{*}{FR $\leftrightarrow$ UK} & WikiMatrix v1 & \multirow{2}{*}{166,063} & \multirow{2}{*}{2,000} & \multirow{2}{*}{2,000} \\
 & \cite{schwenk2019wikimatrix} & & & \\ \hline \hline

 \multicolumn{5}{c}{\textbf{Similar Language Pairs}} \\  \hline 

\multirow{2}{*}{ES $\leftrightarrow$ PT} & TED 2020 v1& \multirow{2}{*}{315,462} & \multirow{2}{*}{2,000} & \multirow{2}{*}{2,000} \\
 & \cite{ted2020} & & & \\ \hline  
 
 \multirow{2}{*}{UK $\leftrightarrow$ RU} & TED 2020 v1& \multirow{2}{*}{197,978} & \multirow{2}{*}{2,000} & \multirow{2}{*}{2,000} \\
 & \cite{ted2020} & & & \\

\Xhline{3\arrayrulewidth}
\end{tabular}
\end{adjustbox}
\caption{Dataset statistics.}
\label{tab:apdx_dataset statistics}
\end{table}

\end{document}